\documentclass[letterpaper]{article} 
\usepackage{aaai25}  
\usepackage{times}  
\usepackage{helvet}  
\usepackage{courier}  
\usepackage[hyphens]{url}  
\usepackage{graphicx} 
\usepackage{natbib}  
\usepackage{caption} 
\usepackage{algorithm}
\usepackage{algorithmic}
\usepackage{wrapfig}
\usepackage{bm}
\usepackage{multirow}

\usepackage{newfloat}
\usepackage{listings}
\DeclareCaptionStyle{ruled}{labelfont=normalfont,labelsep=colon,strut=off} 
\floatstyle{ruled}
\newfloat{listing}{tb}{lst}{}
\floatname{listing}{Listing}
\title{Sharpening Neural Implicit Functions with Frequency Consolidation Priors}
\author{
    Chao Chen\textsuperscript{\rm 1},
    Yu-Shen Liu\textsuperscript{\rm 1}\thanks{The corresponding author is Yu-Shen Liu. This work was supported by National Key R$\&$D Program of China (2022YFC3800600), the National Natural Science Foundation of China (62272263, 62072268), and in part by Tsinghua-Kuaishou Institute of Future Media Data.},
    Zhizhong Han\textsuperscript{\rm 2}
}
\affiliations{
    \textsuperscript{\rm 1}School of Software, Tsinghua University, Beijing, China\\
    \textsuperscript{\rm 2}Department of Computer Science, Wayne State University, Detroit, USA\\

    chenchao19@tsinghua.org.cn, liuyushen@tsinghua.edu.cn, h312h@wayne.edu
}

\usepackage{bibentry}

\begin{document}

\maketitle

\begin{abstract}
Signed Distance Functions (SDFs) are vital implicit representations to represent high fidelity 3D surfaces. Current methods mainly leverage a neural network to learn an SDF from various supervisions including signed distances, 3D point clouds, or multi-view images. However, due to various reasons including the bias of neural network on low frequency content, 3D unaware sampling, sparsity in point clouds, or low resolutions of images, neural implicit representations still struggle to represent geometries with high frequency components like sharp structures, especially for the ones learned from images or point clouds. To overcome this challenge, we introduce a method to sharpen a low frequency SDF observation by recovering its high frequency components, pursuing a sharper and more complete surface. Our key idea is to learn a mapping from a low frequency observation to a full frequency coverage in a data-driven manner, leading to a prior knowledge of shape consolidation in the frequency domain, dubbed frequency consolidation priors. To better generalize a learned prior to unseen shapes, we introduce to represent frequency components as embeddings and disentangle the embedding of the low frequency component from the embedding of the full frequency component. This disentanglement allows the prior to generalize on an unseen low frequency observation by simply recovering its full frequency embedding through a test-time self-reconstruction. Our evaluations under widely used benchmarks or real scenes show that our method can recover high frequency component and produce more accurate surfaces than the latest methods. The code, data, and pre-trained models are available at \url{https://github.com/chenchao15/FCP}.
\end{abstract}

%

\section{Introduction}
\label{sec:intro}

Singed distance Functions (SDFs) can represent high fidelity 3D surfaces with arbitrary topology. An SDF is an implicit function that can predict signed distances at arbitrary 3D query locations. It describes a distance field in the 3D space hosting a surface, where we have iso-surfaces or level sets, each of which has the same signed distance values. One can extract the surface as the zero level set of the SDF using the marching cubes algorithm~\cite{Lorensen87marchingcubes}. 

Recent methods~\cite{mildenhall2020nerf,Oechsle2021ICCV,takikawa2021nglod,takikawa2021nglod} use a neural network to learn an SDF from 3D supervision~\cite{jiang2020lig,Park_2019_CVPR,aminie2022,takikawa2021nglod,Liu2021MLS}, 3D point clouds~\cite{Genova:2019:LST}, or multi-view images~\cite{GEOnEUS2022,Oechsle2021ICCV,neuslingjie,Jiang2019SDFDiffDRcvpr,neuslingjie,Vicini2022sdf,wang2022neuris}, which seamlessly turns a neural network into a neural implicit function. However, due to various reasons like neural networks' bias on low frequency signals, 3D unaware sampling, sparsity in point clouds, or low resolutions of images, neural SDFs still struggle to represent geometries with high frequency components like sharp structures, especially for the ones inferred from point clouds or multi-view images. Although positional encodings~\cite{mildenhall2020nerf} or feature grids~\cite{yu_and_fridovichkeil2021plenoxels,chao2023gridpull} were proposed to recover high frequency components during the inference, their downsides cause either unstable optimization~\cite{sijia2023quantized} or discontinuous representations~\cite{chao2023gridpull}, resulting in either artifacts, or noisy surfaces, holes. Thus, how to recover high frequency components in neural implicit functions is still a challenge.

\begin{figure}[tb]
\vspace{-0.2in}
    \includegraphics[width=0.9\linewidth]{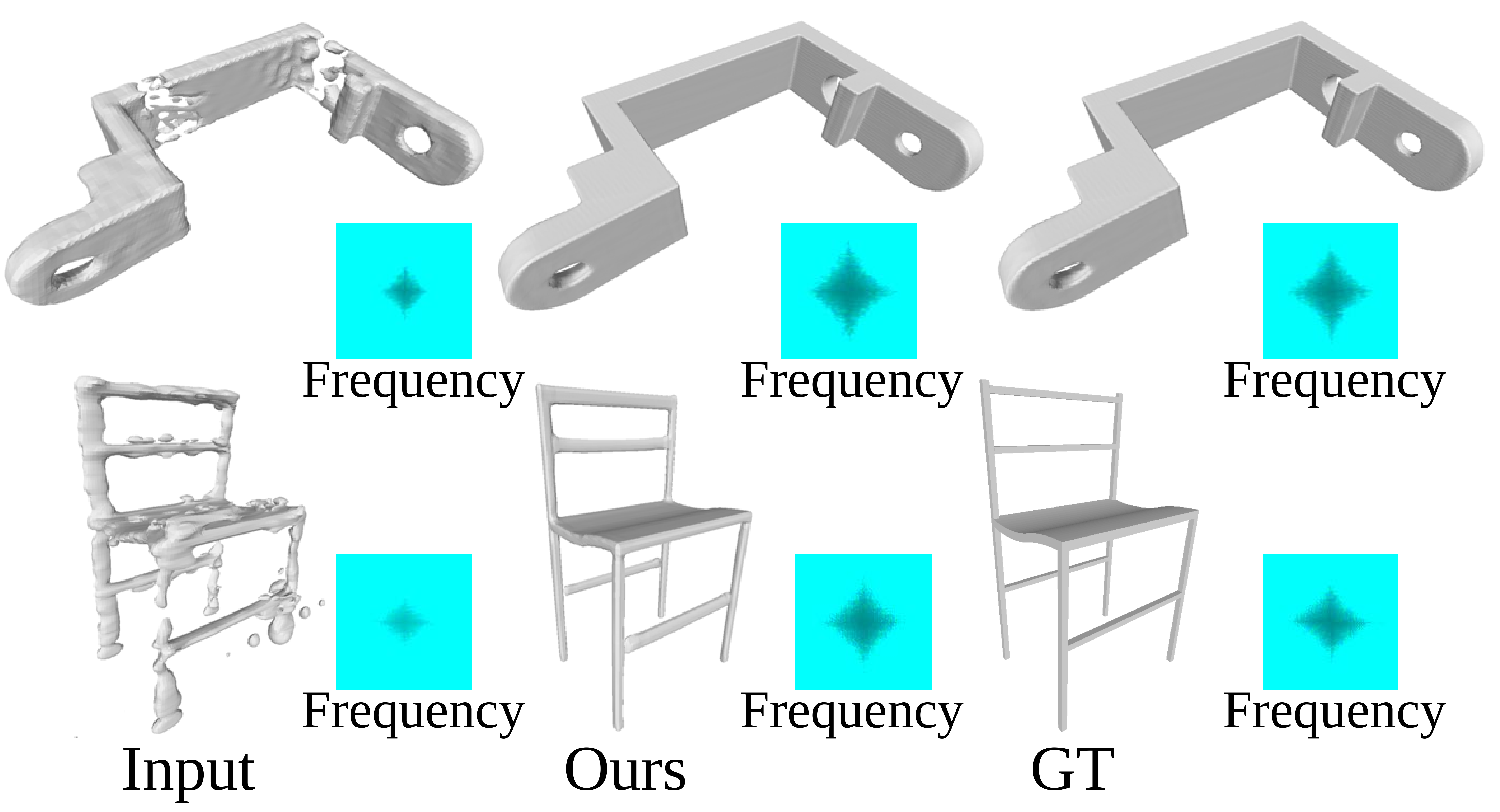}
\vspace{-0.15in}
  \caption{The concept of frequency consolidation priors. We also show averaged frequency weights across a band.}
  \label{fig:frequency}
\vspace{-0.3in}
\end{figure}

To address this challenge, we propose \textit{frequency consolidation priors} to sharpen a neural SDF observation, as illustrated in Fig.~\ref{fig:frequency}. Since sharper features are usually represented by high frequency components, our key idea is to learn a mapping from a low frequency observation to its full frequency coverage in a data-driven manner. The prior knowledge learned by the mapping, dubbed frequency consolidation priors, can produce sharper and more complete surfaces. To generalize a learned prior on unobserved low frequency SDFs better, we introduce to represent frequency components as embeddings, and disentangle the embedding of low frequency components from the one of its full frequency coverage. Our design enables the learned prior to recover full frequency embeddings by overfitting unseen low-frequency observations through a test-time self-reconstruction. We learn a frequency consolidation prior by establishing a dataset containing low and full frequency component pairs, where we produce low frequent components by removing high frequencies from full frequency coverage of a shape in the frequency domain. We demonstrate the effectiveness and the good generalization of our prior in shape and scene modeling. Benchmark comparisons show our method's superiority in accuracy and generalization over the latest methods. Our contributions are listed below.

\begin{itemize}
    \item We present a novel method to sharpen neural SDFs for sharper and more complete surfaces in the frequency domain. Our frequency consolidation prior can recover full frequency coverage from a low frequency observation.
    \item We justify the idea of representing frequency components as embeddings. This design can prompt the generalization of learned priors by recovering the embedding of full frequency coverage via a test-time self-reconstruction on low frequency observations.
    \item We report the state-of-the-art results in shape or scene modeling by sharpening reconstructions from sparse point clouds or multi-view images.
\end{itemize}

\section{Related Work}
\noindent\textbf{Neural Implicit Representations. }Neural implicit representations have made huge progress in representing 3D geometry~\cite{MeschederNetworks,Jiang2019SDFDiffDRcvpr,chaompi2022,Hu2023LNI-ADFP,sijia2023quantized,localn2nm2024}. One can learn neural implicit representations using coordinate-based MLP from supervision including 3D ground truth distances~\cite{jiang2020lig,DBLP:conf/eccv/ChabraLISSLN20,Peng2020ECCV,takikawa2021nglod,Liu2021MLS}, 3D point clouds~\cite{Zhizhong2021icml,chaompi2022,neuraltpstpami,multipull2024,n2n_tpami,capudf_tpami}, or multi-view images~\cite{mildenhall2020nerf,GEOnEUS2022,Oechsle2021ICCV,neuslingjie,Vicini2022sdf,wang2022neuris,guo2022manhattan,sijia2023quantized}. 
With differentiable renderers, neural implicit representations can be learned by minimizing errors between their 2D renderings and ground truth images. Using surface rendering~\cite{Jiang2019SDFDiffDRcvpr}, DVR~\cite{DVRcvpr} and IDR~\cite{yariv2020multiview} estimate geometry in a radiance field. IDR also models view direction as a condition to reconstruct high frequency details. Since these methods focus on intersections on surfaces, they need masks to filter out the background.

NeRF~\cite{mildenhall2020nerf} and its variations~\cite{yu_and_fridovichkeil2021plenoxels,zhang2024learning} simultaneously model geometry and color using volume rendering. They aim to generate novel views, and render images without masks. By deriving novel rendering equations, UNISURF~\cite{Oechsle2021ICCV} and NeuS~\cite{neuslingjie} are able to render occupancy and signed distance fields into RGB images, which measures the errors of implicit functions. Following methods improve accuracy of implicit functions using priors or losses including depth~\cite{wang2023coslam,Hu2023LNI-ADFP}, normals~\cite{wang2022neuris,guo2022manhattan}, multi-view consistency~\cite{GEOnEUS2022}, and segmentation priors~\cite{haghighi2023neural}.

\noindent\textbf{Learning with Frequency. }Learning neural implicit representations with multi-scale details enhances interpretability. It allows progressively detailed visualization at different scales~\cite{takikawa2021nglod}. Controlling curvature regulation~\cite{ehret2022regularization} can add or remove surface details. A common approach is learning neural implicits with several frequency bands that cover a whole frequency scope~\cite{lindell2021bacon, grattarola2022generalised}. This yields a multi-scale representation by reconstructing surfaces or signals from different frequency bands. Moreover, SAP learns an occupancy function by solving a Poisson equation in the frequency domain~\cite{Peng2021SAP}.

\begin{figure*}[t]
  \centering
    \vspace{-0.5cm}
  \includegraphics[width=0.92\linewidth]{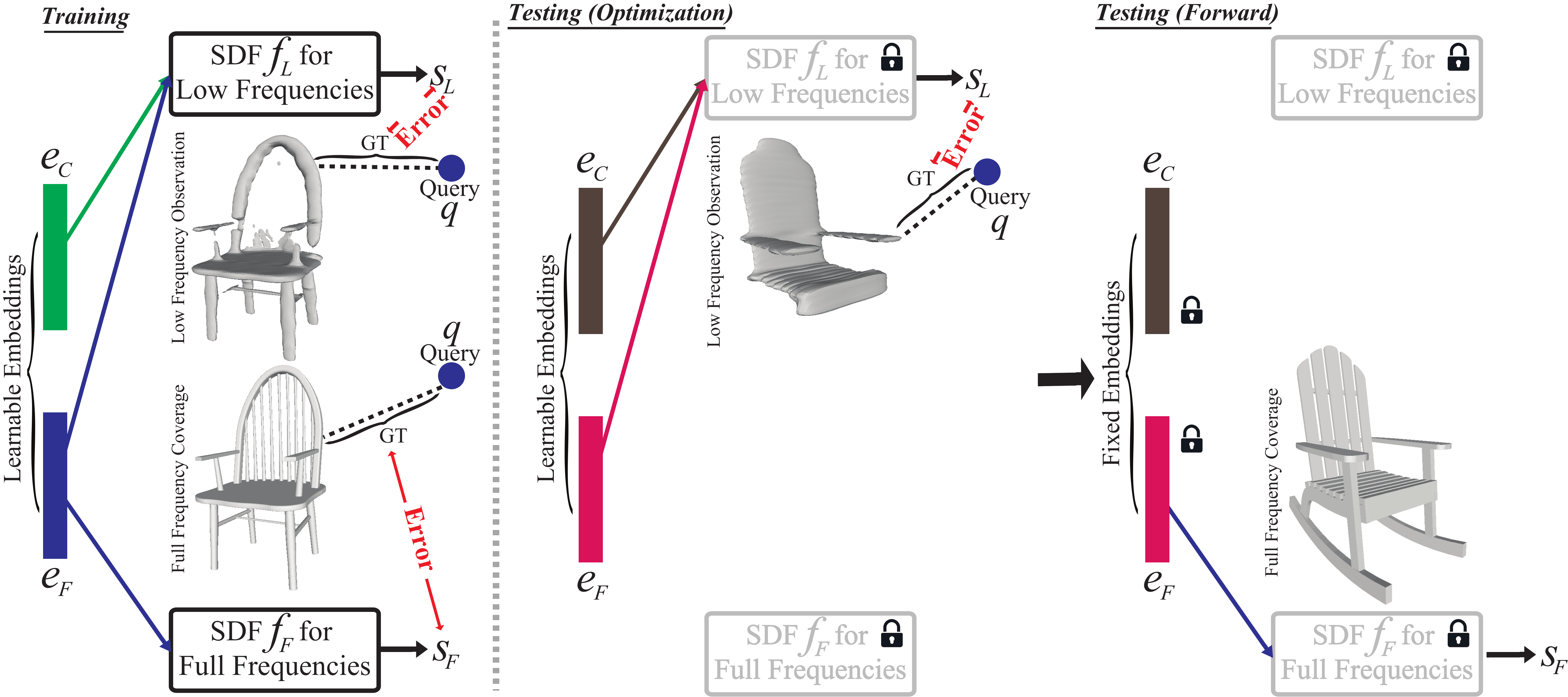}
  \vspace{-0.2cm}
  \caption{The overview of our method.} 
  \vspace{-0.5cm}
  \label{fig:overview}
\end{figure*}

\noindent\textbf{Recovering Sharp Structures. }There are several different strategies to recover sharp edges. Dual contouring~\cite{chen2022ndc} can reconstruct sharper edges than the marching cubes~\cite{Lorensen87marchingcubes} with the help of gradients. By modeling displacements, more high frequency details can get recovered on surfaces~\cite{yifan2021geometry}. Edges are also an important structure to recover, especially in CAD modeling. NEF~\cite{Ye_2023_CVPR} was proposed to learn an implicit function to represent edges from multi-view images. Some methods~\cite{Lambourne_2022} focus on sharpening edges directly on 3D shapes. However, they merely work on relatively clean and plausible shapes. Consolidation is also a way to sharpen shapes, especially for point clouds \cite{metzer2020self}, which can generate points with sharp features or in sparse regions, and also remove noises and outliers.

Unlike previous methods, we aim to sharpen implicit field poorly recovered from point clouds or multi-views. We use a data-driven strategy to sharp a shape by learning priors in the frequency domain, which consolidates the frequency components and completes missing structures as well.

\section{Method}
\label{sec:method}
\noindent\textbf{Overview. }Our method aims to sharpen a low frequency observation represented by an SDF $f_L$ (or a point cloud), as illustrated in Fig.~\ref{fig:overview}. With $f_L$, we intend to recover its full frequency coverage as another SDF $f_F$ which represents a surface with sharper and more complete structures than the low frequency observation. Both $f_L$ and $f_F$ are learned by neural networks with parameters $\bm{\theta}_L$ and $\bm{\theta}_F$, respectively. At an arbitrary query $q$, $f_L$ and $f_F$ predict signed distances as $s_L=f_L(q,\bm{e}_L)$ and $s_F=f_F(q,\bm{e}_F)$, respectively, where $\bm{e}_L$ and $\bm{e}_F$ are learnable embeddings representing low frequency components and full frequency coverage (or shape identities), respectively, which are also conditions in $f_L$ and $f_F$. We use $\bm{e}_L$ and $\bm{e}_F$ as bridges to connect $f_L$ and $f_F$, where $\bm{e}_L$ is formed by $\bm{e}_C$ and $\bm{e}_F$.

\begin{figure}[tb]
 \vspace{-0.1in}
    \includegraphics[width=\linewidth]{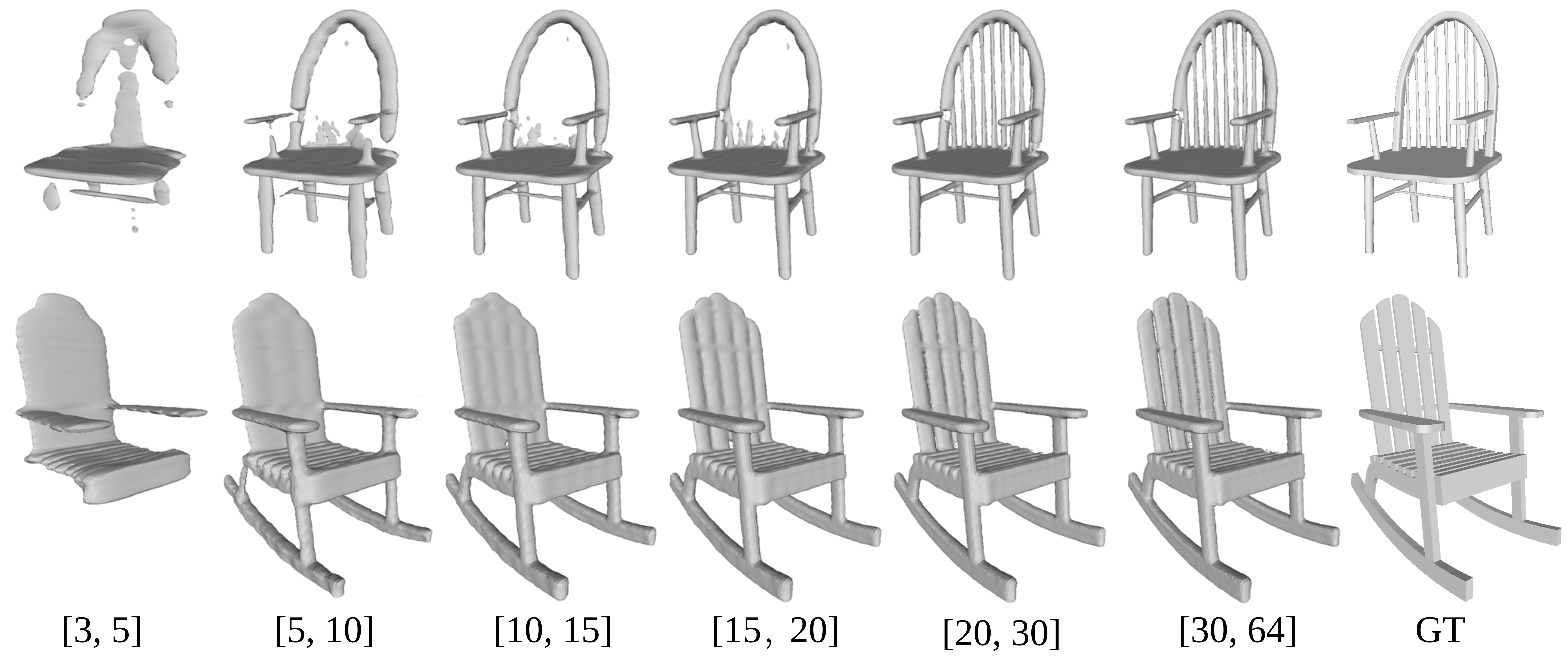}
    \vspace{-0.25in}
  \caption{The illustration of low frequency observations and the full frequency coverage.}
  \label{fig:supervisions}
  \vspace{-0.25in}
\end{figure}

We learn a frequency consolidation prior by learning $f_L$ and $f_F$ in a data-driven manner using supervision established from ground truth meshes. During testing, given an unseen SDF with low frequency components, we generalize the learned prior by conducting a test-time self-reconstruction, which learns the embeddings $\bm{e}_L$ and $\bm{e}_F$ of the shape using $f_L$ with the fixed parameters $\bm{\theta}_L$. Then, we further sharpen the shape by decoding the learned $\bm{e}_F$ using $f_F$ with the fixed parameters $\bm{\theta}_F$.

\begin{wrapfigure}{r}{.5\linewidth}
    \vspace{-0.25in}
    \includegraphics[width=\linewidth]{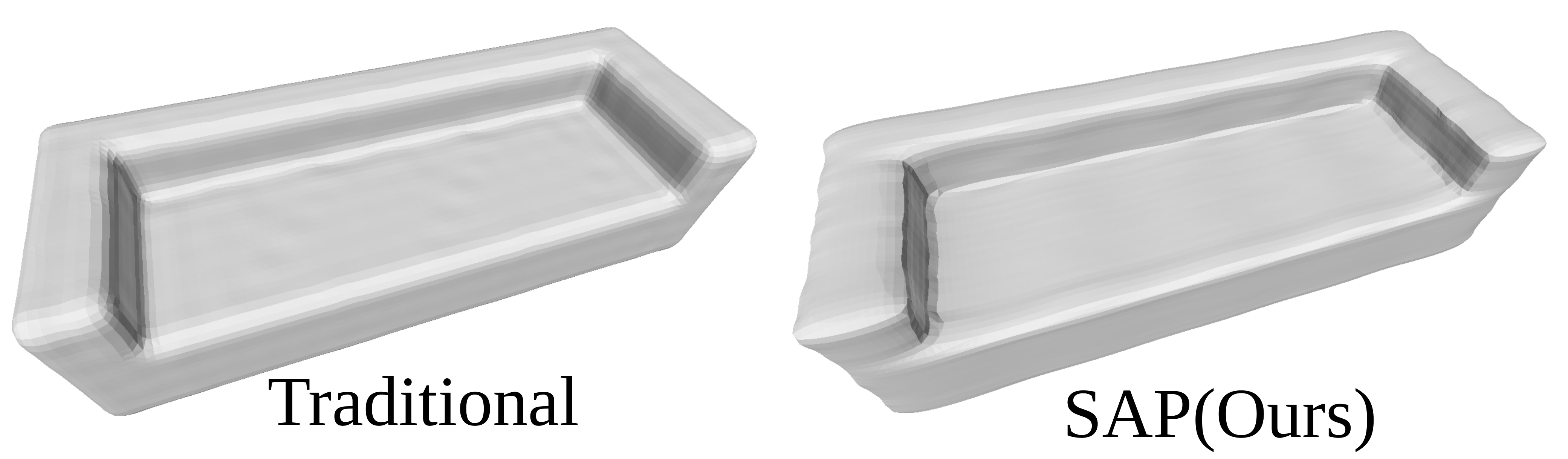}
  \vspace{-0.25in}
  \caption{Over-smoothed surfaces.}
  \label{fig:spectral}
  \vspace{-0.22in}
\end{wrapfigure}

\noindent\textbf{Supervisions for Learning Priors. }We establish supervisions from ground truth meshes. For a shape $M$, we produce its low frequency observations $M_L$ by randomly removing its high frequency components from its full frequency coverage $M_F$. To decompose a 3D mesh into frequency domain, the traditional method like the spectral geometry theory~\cite{zhang_eg07} does eigen-decomposition of the discrete Laplace–Beltrami operator and regards the eigenvectors as frequency components. Some learning-based methods~\cite{lindell2021bacon,takikawa2021nglod} are also alternatives. However, eigen-decomposing a large matrix whose dimension is determined by the vertex number is usually limited due to the large space complexity, while learning based methods are too slow to get enough samples as supervisions. Instead, we introduce to manipulate frequency components in solving Poisson surface reconstruction from point clouds for efficiency. As a fast solving PDE strategy, spectral methods solve a Poisson surface reconstruction problem using Fast Fourier Transform (FFT)~\cite{Peng2021SAP}. The low frequency observations (SDF) established by our method with proper low frequency band could produce over smoothed surfaces which are very similar to the one produced by spectral geometry theory in Fig.~\ref{fig:spectral}.

Specifically, we first randomly sample dense points on the mesh of $M$, and then estimate an occupancy field by solving a Poisson surface reconstruction equation, where we obtain magnitudes for frequencies in a frequency band. We reconstruct $M$ by running the marching cubes algorithms~\cite{Lorensen87marchingcubes} with the estimated occupancy function, and use the reconstruction as the full frequency coverage $M_F$. At the same time, we produce low frequency occupancy functions by removing some parts of high frequency components (setting their corresponding magnitudes to 0), and use each one of the manipulated occupancy functions to reconstruct a mesh as a low frequency observation $M_L$. To make low frequency observations cover all frequencies over the frequency band $[0,64]$ illustrated in Fig.~\ref{fig:supervisions}, we randomly select one frequency from one of six subbands, such as ``[3,5]'' or ``[5,10]'', and remove all frequencies larger than the sampled one. For instance, we select 4 in ``[3,5]'', then we will get a low frequency observation $M_L$ by removing all frequencies from $5$ to $64$. Each low frequency observation produced is paired with the full frequency coverage $M_F$ as a training sample. Fig.~\ref{fig:supervisions} illustrates $6$ pairs of low frequency observations $\{M_L\}$ and their corresponding full frequency coverage $M_F$ (rightmost) on two shapes. Note that we do not simply use the ground truth meshes as a full frequency coverage $M_F$ to avoid non-watertight meshes during training. 

Moreover, we produce more low frequency observations in the band $[3,30]$, as most reconstructions in real applications contain very low-frequency components. These observations not only have smooth surfaces but also severe structural corruptions. Using them as training samples enables our prior to handle extremely poor reconstructions well.

\noindent\textbf{Frequency Component Modeling. }With low frequency observation $M_L$ and its corresponding full frequency coverage $M_F$, we learn a frequency consolidation prior as a mapping from $M_L$ to $M_F$ in Fig.~\ref{fig:overview}. We represent both $M_L$ to $M_F$ as SDFs $f_L$ and $f_F$ which are approximated by a two-branch network parameterized by $\bm{\theta_L}$ and $\bm{\theta_F}$. At each query $q$, one branch predicts a signed distance $s_L=f_L(q,\bm{e}_L)$ around $M_L$, the other predicts a signed distance $s_F=f_F(q,\bm{e}_F)$ around $M_F$. We use embeddings $\bm{e}_L$ and $\bm{e}_F$ to model $M_L$ and $M_F$, which are also used as conditions to distinguish the low frequency band and shape identity when sharing the same neural network implementation.

For an embedding $\bm{e}_L$, we formulate it as a learnable $256$-dimensional vector, and assign it to a low frequency observation $M_L$. Similarly, we formulate an embedding $\bm{e}_F$ as a learnable $128$-dimensional vector, assign it to the full frequency coverage $M_F$, and more importantly, make it shareable to all low frequency observations $\{M_L\}$ of shape $M$. 

We bridge the two neural SDFs $f_L$ and $f_F$ by disentangling $\bm{e}_F$ from $\bm{e}_L$. We formulate $\bm{e}_L$ as a concatenation of $\bm{e}_F$ representing a shape identity $M$ and $\bm{e}_C$ representing a frequency corruption on the specific $M_F$ below,

\vspace{-0.1in}
\begin{equation}
\label{eq:embedding}
\bm{e}_L=[\bm{e}_F \ \ \bm{e}_C].
\end{equation}

This disentangling makes the frequency modeling interpretable, compacts the embedding space, synchronizes the learning of $f_L$ and $f_F$, and more importantly, increases the generalization ability of the learned frequency consolidation prior which will show in experiments.


\noindent\textbf{Learning Frequency Consolidation Priors. }To learn the prior, we train the two-branch network to regress signed distances at query $q$. With a low frequency observation $M_L$ and its target $M_F$, we sample queries $q$ around $M_F$ and record the ground truth signed distances $s_L^{gt}$ and $s_F^{gt}$. We optimize parameters by minimizing the prediction error denoted by
%
\begin{equation}
\label{eq:mse}
\min_{\bm{\theta}_L,\bm{\theta}_F,\{\bm{e}_F\},\{\bm{e}_C\}} ||s_L-s_L^{gt}||_2^2+||s_F-s_F^{gt}||_2^2,
\end{equation}
%
\noindent where $s_L=f_L(q,[\bm{e}_F \ \ \bm{e}_C])$ and $s_F=f_F(q,\bm{e}_F)$ are signed distance predictions. 

\noindent\textbf{Generalizing Frequency Consolidation Priors. }We generalize the learned prior to sharpen an unseen low frequency observation $M_L'$. $M_L'$ can be represented as an SDF, a point cloud, or a mesh. To leverage the learned prior, we transform a point cloud or a mesh into an SDF using surface reconstruction methods like NeuralPull~\cite{Zhizhong2021icml}.

With our disentangling of $\bm{e}_F$ from $\bm{e}_L$, we can estimate the shape identity $\bm{e}_F$ through a test-time optimization in self-reconstruction on $M_L'$ as auto-decoding~\cite{Park_2019_CVPR}. To this end, we sample queries $q$ around $M_L'$ and record signed distances $s_L^{gt}$' as supervision. We estimate $\bm{e}'_L=[\bm{e}_F' \ \ \bm{e}_C']$ with fixed parameters $\bm{\theta}_L$ by minimizing the reconstruction errors below,

\vspace{-0.1in}
\begin{equation}
\label{eq:mse}
\min_{\bm{e}_F',\bm{e}_C'} ||s_L-{s_L^{gt}}'||_2^2.
\end{equation}

After the optimization, we represent the SDF of a full frequency coverage $M_F'$ as $f_F(q,\bm{e}_F')$. We can reconstruct the surface of $M_F'$ by running the marching cubes~\cite{Lorensen87marchingcubes} with $f_F(q,\bm{e}_F')$.

\noindent\textbf{Implementation Details. } 
We adopt two Gaussian functions centered at each point with standard deviations $\sigma_{1}$ and $\sigma_{2}$ to sample queries. Starting from meshes, we sample dense point clouds as surface points, and sample queries around each surface point. We set $\sigma_{1}$ to 8 for full-space sampling, allowing the network to perceive a large space and cover various shape variations. $\sigma_{2}$ is set to 0.2, enabling queries to be sampled close to the surface. These two types of queries are sampled with a one-to-one weighting ratio, dynamically sampling 16,384 queries in each iteration.

We learn $\bm{e}_L$ and $\bm{e}_F$ by 3 fully connected layers with 128 hidden units and a ReLU on each layer. We employ two SDF-decoder networks similar to DeepSDF~\cite{Park_2019_CVPR} to learn $f_L$ and $f_F$.  The Adam optimizer is used with an initial embedding learning rate of 0.0005 and an SDF-decoder learning rate of 0.001, both decreased by 0.5 every 500 epochs. We train our model in 2000 epochs. During test-time optimization, we overfit $f_L$ on a low frequency observation in 800 iterations with a learning rate of 0.005.

\begin{table}[htb]
\vspace{-0.15in}
\centering
\resizebox{\linewidth}{!}{
    \begin{tabular}{c|c|c|c}  
     \hline
     \multicolumn{1}{c|}{Method}&$CD_{L1}\times10$&$CD_{L2}\times100$&$NC$\\
     \hline
     DeepSDF~\cite{Park_2019_CVPR}&0.287&0.381&0.804\\
     ConvOcc~\cite{Peng2020ECCV}&0.306&0.451&0.805\\
     LIG~\cite{jiang2020lig}&0.292&0.430&0.809\\
     IDF~\cite{yifan2021geometry}&0.287&0.390&0.815\\
     NDC~\cite{chen2022ndc}&0.269&0.358&0.768\\
     POCO(pretrained)&0.259&0.374&0.812\\
     POCO~\cite{pococvpr2022}&0.217&0.284&0.858\\
     ALTO(pretrained)&0.253&0.367&0.819\\
     ALTO~\cite{wang2022alto}&0.213&0.285&0.861\\
     \hline
     Ours&\textbf{0.187}&\textbf{0.216}&\textbf{0.871}\\
     \hline
   \end{tabular}}
   \vspace{-0.14in}
   \caption{Reconstruction accuracy of 13 classes on ShapeNet in terms of $CD_{L1}$, $CD_{L2}$ and $NC$. The accuracy for each class is provided in the supplement.}  
   \label{table:NOX31}
   \vspace{-0.25in}
\end{table}

\begin{figure*}[t]
  \centering
   \vspace{-0.0in}
  \includegraphics[width=\linewidth]{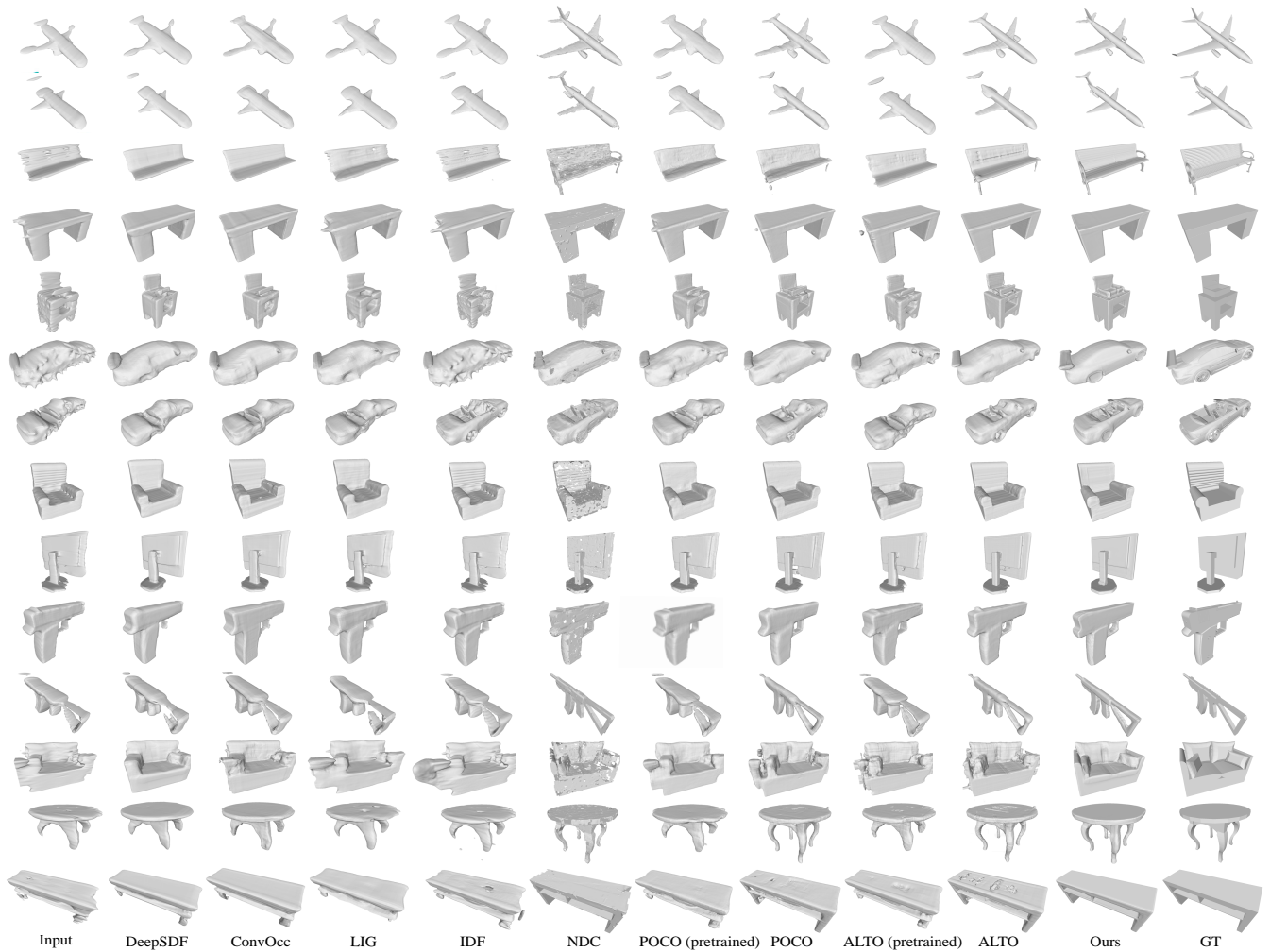}
   \vspace{-0.1in}
  \caption{Visual comparison with the state-of-the-art on ShapeNet.}
  \vspace{-0.25in}
  \label{fig:shapenet}
\end{figure*}

\section{Experiments and Anaylysis}
\noindent\textbf{Datasets and Metrics. }
We evaluate our method by numerical and visual comparisons with the latest methods on ShapeNet~\cite{shapenet2015}, ABC~\cite{Koch_2019_CVPR}, and ScanNet datasets~\cite{dai2017scannet}. For ShapeNet, we report evaluations under 13 classes with the train/test split from 3D-R2N2 \cite{ChoyXGCS16} and also on 8 classes under the test split from NeuralTPS \cite{NeuralTPS} for fair comparisons. For ABC, we follow Points2Surf~\cite{ErlerEtAl:Points2Surf:ECCV:2020}, and use its train/test splitting for evaluation. For ScanNet, we follow Neural Part Priors \cite{bokhovkin2022neuralparts} and use the same 6 classes and test split. All experiments leverage marching cubes \cite{Lorensen87marchingcubes} on a $256^3$ grid to reconstruct meshes.

For the ShapeNet dataset, we measure errors using L1 Chamfer Distance ($CD_{L1}$), L2 Chamfer Distance ($CD_{L2}$), and normal consistency ($NC$). Following NeuralTPS \cite{NeuralTPS}, we randomly sample 100k points on the reconstructed and ground truth meshes. Under the ScanNet dataset, we follow Neural Part Priors to report L1 Chamfer Distance ($CD_{L1}$) between reconstructed meshes and ground truth meshes transformed to the ScanNet coordinate space. Notice that, in Neural Parts Prior, each shape is evaluated over the union of the $CD_{L1}$ of predicted and ground truth parts, which is equivalent to the $CD_{L1}$ between global predictions and global ground truths. We report two results using annotated shapes from Scan2CAD \cite{Avetisyan_2019_CVPR} dataset and extract shapes using ScanNet segmentation masks as ground truth, respectively.

\noindent\textbf{Learning Frequency Consolidation Priors. }For each shape used for training, we use the first $5$ low frequency observations (as shown in Fig.~\ref{fig:supervisions}) and the full frequency coverage to learn the frequency consolidation prior. We sample queries around both the low frequency observations and the full frequency coverage, and record the signed distances to both of the low and full frequency meshes as supervision.
 
\subsection{Evaluations}
\noindent\textbf{Evaluation on ShapeNets. }
We evaluate using frequency consolidation priors learned from training samples. For each test shape, we generate low-frequency observations as described and use the worst observation to assess all methods. Tab.~\ref{table:NOX31} reports average evaluations across all classes, showing our method achieves the best performance. Detailed per-class results are in the supplementary materials.

We use pre-trained parameters or parameters retrained using our data to produce the results of the latest methods. To train DeepSdf~\cite{Park_2019_CVPR}, POCO~\cite{pococvpr2022}, and ALTO~\cite{wang2022alto} which map a point cloud into an SDF, we sample the low frequency observation as a point cloud, record signed distances at queries near their full frequency coverages, forming training samples for both POCO and ALTO. Our method significantly outperforms these methods. Visual comparisons in Fig.~\ref{fig:shapenet} show that DeepSDF, POCO and ALTO struggle to generalize their prior knowledge on various low-frequency observations. With pre-trained parameters, NDC~\cite{chen2022ndc} produces sharp edges with dual contouring, preserving more and sharper structures than the marching cubes algorithm ~\cite{Lorensen87marchingcubes}. But NDC does not generalize well on low frequency shapes and produces broke and noisy surfaces, as shown in Fig.~\ref{fig:shapenet}. IDF~\cite{yifan2021geometry} and ConvOcc~\cite{Peng2020ECCV} pretrained on corrupted shapes can also generate high frequency geometry on surfaces from corrupted point clouds in the spatial space. Comparisons in Tab.~\ref{table:NOX31} and Fig.~\ref{fig:shapenet} show that it can not handle large geometry variations. 

\begin{figure}[t]
  \centering
  \vspace{-0.0in}
  \includegraphics[width=\linewidth]{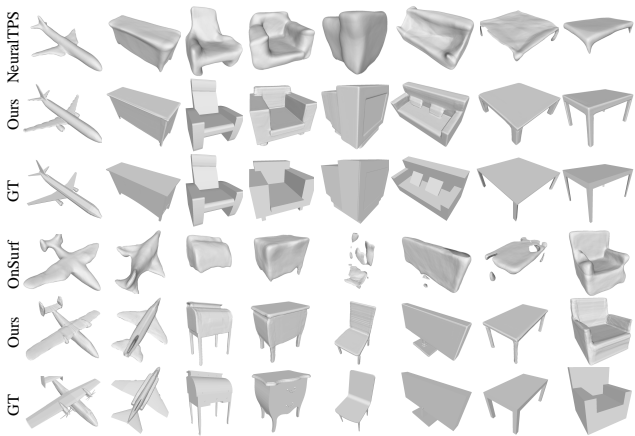}
  \vspace{-0.1in}
  \caption{Reconstruction results for test-time optimization with sparse reconstructions from NeuralTPS and OnSurf as low-frequency observations, respectively.}
  \vspace{-0.25in}
  \label{fig:tps}
\end{figure}

\noindent\textbf{Refining Reconstructions from Sparse Point clouds. }We further evaluate the generalization ability of our learned prior. In the previous experiment, we produce test shapes from a known frequency band used in training. How good the performance of our prior is on unobserved frequency bands will be evaluated in this experiment. We use reconstructions from sparse point clouds by the latest methods as test shapes, which are barely with any geometry details as shown in Fig.~\ref{fig:tps} and have unobserved frequency components. 

\begin{table}[b]
\vspace{-0.15in}
\centering
\resizebox{0.9\linewidth}{!}{
    \begin{tabular}{c|c|c|c}  
     \hline
     Method&$CD_{L1}\times10$&$CD_{L2}\times100$&$NC$\\
     \hline
     Onsurf~\cite{onsurfacepriors2022}&0.214&0.223&0.845\\
     Onsurf+Ours&\textbf{0.180}&\textbf{0.165}&\textbf{0.886}\\
    \hline
     NeuralTPS~\cite{NeuralTPS}&0.141&0.093&50.899\\
     NeuralTPS+Ours&\textbf{0.115}&\textbf{0.088}&\textbf{0.918}\\
     \hline
   \end{tabular}}
   \vspace{-0.1in}
   \caption{Numerical comparisons with sparse point cloud reconstruction on ShapeNet.}  
   \label{table:NOX31tps}
   \vspace{-0.35in}
\end{table}

\begin{wrapfigure}{r}{.3\linewidth}
\vspace{-0.15in}
    \includegraphics[width=\linewidth]{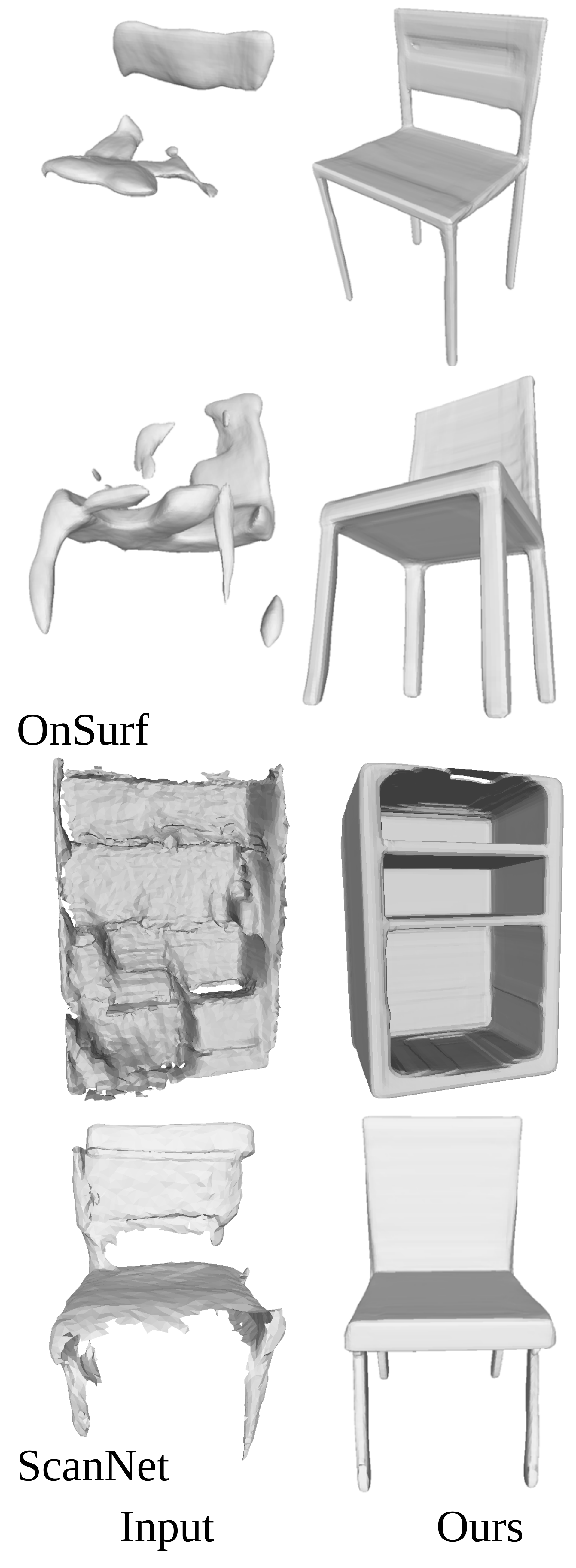}
\vspace{-0.25in}
  \caption{Reconstrcutions from severely corrupted cases.}
  \label{fig:badcase}
\vspace{-0.15in}
\end{wrapfigure}

We use NeuralTPS~\cite{NeuralTPS} and OnSurf Prior~\cite{onsurfacepriors2022}, state-of-the-art methods for sparse point cloud reconstruction, to generate test shapes from 300 points across 8 ShapeNet classes. Using our learned prior, we recover high-frequency components. Tab.\ref{table:NOX31tps}, Fig.\ref{fig:tps}, and Fig.~\ref{fig:badcase} (first two rows) show our method generalizes well to unknown frequency bands, improving reconstruction accuracy with sharper edges, flatter planes, and more complete surfaces. Detailed per-class evaluations are in the supplementary materials.

\begin{figure}[t]
  \centering
  \vspace{-0.0in}
  \includegraphics[width=\linewidth]{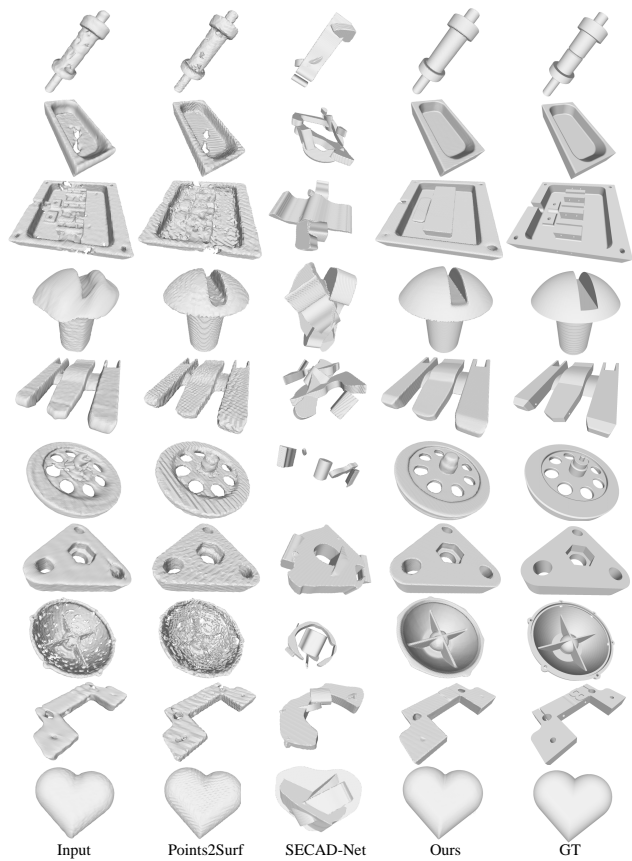}
  \vspace{-0.15in}
  \caption{Visual comparisons on ABC.}
  \vspace{-0.3in}
  \label{fig:supervisionsabc}
\end{figure}

\noindent\textbf{Evaluations on CAD Modeling. }CAD shapes usually contain many sharper edges. We learn a prior from shapes in the training dataset collected from ABC dataset~\cite{Koch_2019_CVPR} by Points2Surf~\cite{ErlerEtAl:Points2Surf:ECCV:2020}. We use the same way to produce training samples using $5$ low frequency and a full frequency coverage from each shape for training. For each testing shape, we also produce $5$ low frequency observations, and use each one as a testing sample. We calculate the mean and variance of the $5$ evaluations for each testing shape. Tab.~\ref{table:NOX31abc} reports the average values over all testing shapes, comparing our method with Points2Surf~\cite{ErlerEtAl:Points2Surf:ECCV:2020} and SECAD-Net~\cite{li2023secadnet}. Our method achieves the best reconstruction accuracy and stability. Visual comparisons in Fig.~\ref{fig:supervisionsabc} show that our method can generate much sharper edges and more accurate structures. Points2Surf struggles with surface variations, and SECAD-Net generates sharper edges but does not generalize well.

\begin{table}[b]
\vspace{-0.2in}
\centering
\resizebox{0.9\linewidth}{!}{
    \begin{tabular}{c|c|c|c|c}  
     \hline
     \multirow{2}{*}{Method}&\multicolumn{2}{c|}{$CD_{L1}$}&\multicolumn{2}{c}{$NC$}\\
     \cline{2-5}
     &Mean&Variance&Mean&Variance\\
     \hline
     Points2Surf~\cite{ErlerEtAl:Points2Surf:ECCV:2020}&0.014&0.431&0.902&5.166\\
     SECAD-Net~\cite{li2023secadnet}&0.041&0.178&0.800&1.370\\
     \hline
     Ours&\textbf{0.011}&\textbf{0.015}&\textbf{0.962}&\textbf{1.076}\\
     \hline
   \end{tabular}}
   \vspace{-0.1in}
   \caption{Accuracy of reconstruction on ABC dataset in terms of $CD_{L1}$ and $NC$. We multiply both variances by $10^{4}$.}
   \label{table:NOX31abc}
   \vspace{-0.15in}
\end{table}

\begin{figure}[t]
  \centering
  \vspace{-0.0in}
  \includegraphics[width=\linewidth]{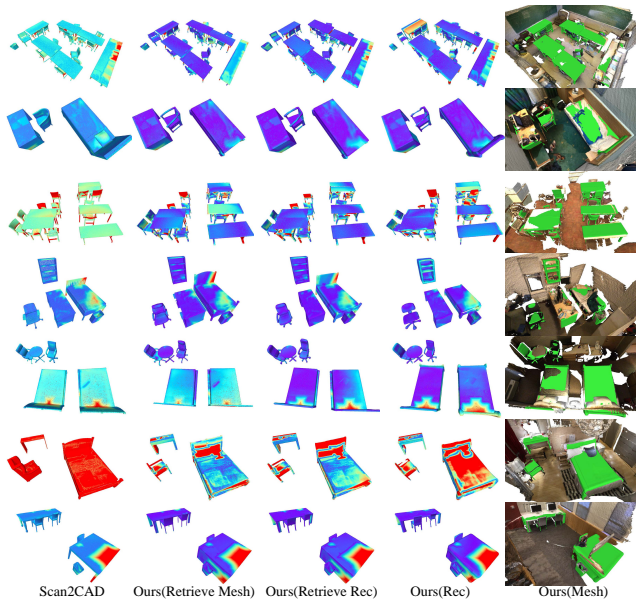}
  \vspace{-0.2in}
  \caption{Visual comparisons on ScanNet dataset. The red in error maps indicates larger errors.}
  \vspace{-0.25in}
  \label{fig:supervisionsscannet}
\end{figure}

\noindent\textbf{Reconstruction in Scenes. }Our learned prior also works with objects in scene modeling. We learn our priors on ShapeNet classes that appear in the scenes used for evaluation. These scenes are reconstructed from real scans. We use the GT segmentation masks to segment shapes as partial meshes. These partial meshes are also with artifacts, few geometry, unobserved frequency bands or severe corruption. We use the poses and scale information from Scan2CAD~\cite{Avetisyan_2019_CVPR} to determine the layout in our visualizations.

We use two kinds of GT shapes in evaluations in Tab. 1 and Tab. 2 in our supplement, respectively. One is the shapes provided by Scan2CAD, which are retrieved from ShapeNet. These shapes are complete but may drift away a lot from the real scans. The other kind is the shapes obtained directly from real scans with GT segmentation masks. These shapes are mostly incomplete but more identical to the real scenes. 

With a shape segmented from a scene, we use NeuralPull to reconstruct a coarse but watertight mesh, serving as a low frequency observation, We then generalize the learned prior to recover its full frequency coverage (Rec). Using the low frequency observation, we can also produce two results by retrieving all low frequency observations in our training set. Specifically, we render $30$ images from viewpoints around each shape. We use clip image encoder~\cite{DBLP:journals/corr/abs-2103-00020} to extract features of each image. Then, we use a single direction CD distance as a retrieval metric to evaluate the distance between two sets of images representing two shapes. 

For each retrieved low frequency observation, we use the reconstructed high frequency coverage or their GT meshes from ShapeNet to report the results including (Retrieve Rec) and (Retrieval Mesh). We report the evaluations of these three results in both Tab. 1 and Tab. 2 in our supplement.



Our reconstruction results produce more accurate reconstruction than NeuralPartPriors~\cite{bokhovkin2022neuralparts} and PartUnderstanding~\cite{TPBU2021} using Scan2CAD as GT shapes. Based on that, our retrieved results can produce even better results. Similarly, comparisons in Tab. 2 in our supplement show our superiority over Scan2CAD with segmented meshes as ground truth shapes. We detail our results with error maps in Fig.~\ref{fig:supervisionsscannet}, and show plausible results on bad reconstructions from scenes in ScanNet in Fig.~\ref{fig:badcase} (last two rows). We can see that shapes retrieved by Scan2CAD are not every identical to the real scans.


\begin{figure}[t]
  \centering
  \vspace{-0.0in}
  \includegraphics[width=\linewidth]{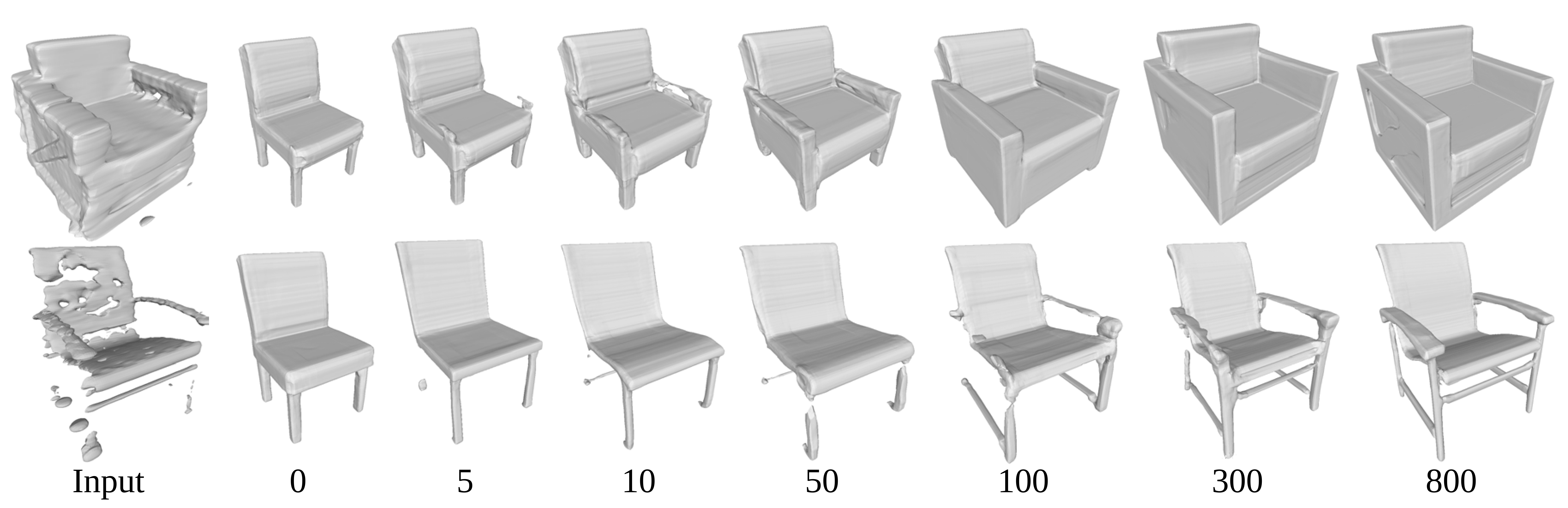}
  \vspace{-0.5cm}
  \caption{Visualization of the test-time optimization.}
  \vspace{-0.1in}
  \label{fig:optim}
\end{figure}

\begin{figure}[t]
  \centering
  \includegraphics[width=\linewidth]{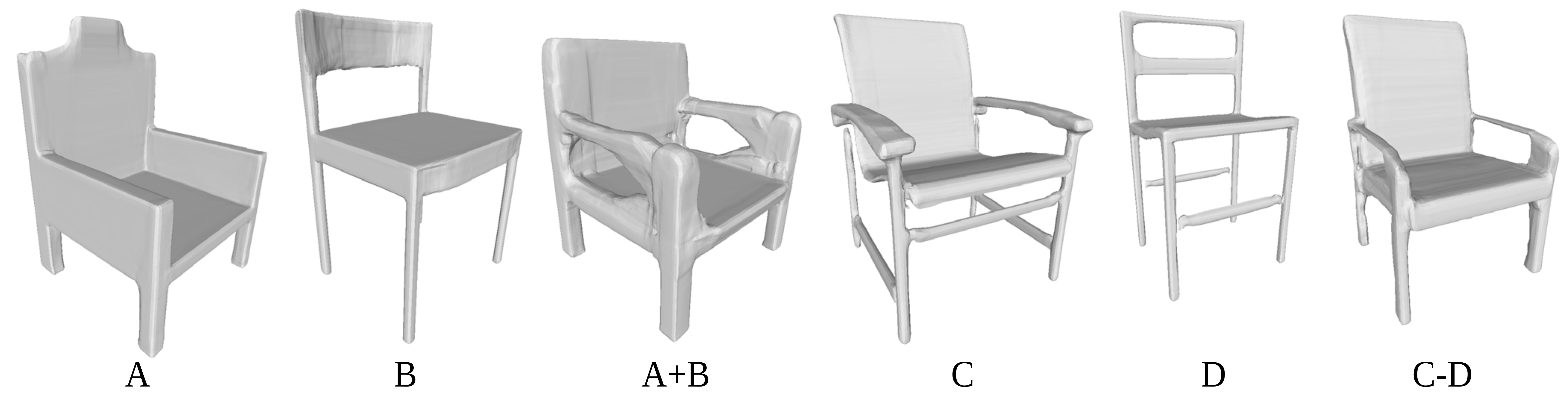}
  \vspace{-0.4cm}
  \caption{Embedding manipulations for shape generation.}
  \vspace{-0.2in}
  \label{fig:optimtadd}
\end{figure}


\subsection{Ablation Studies and Analysis}

\begin{figure}[h]
\vspace{-0.15in}
    \includegraphics[width=\linewidth]{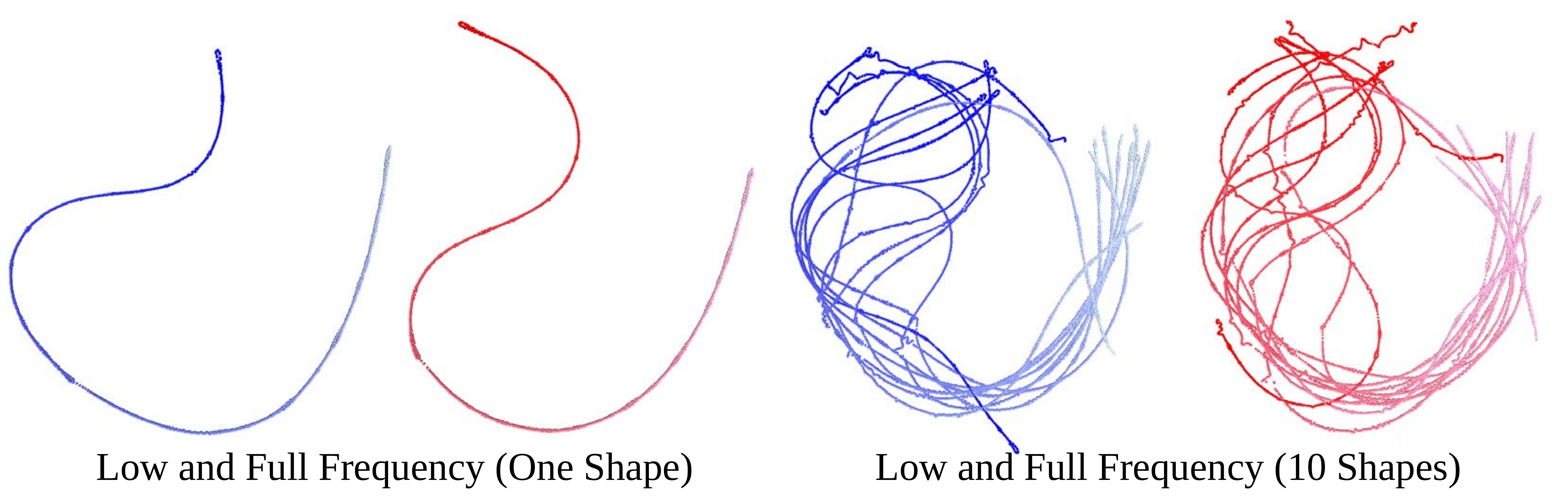}
\vspace{-0.25in}
  \caption{Embeddings for low and full frequency shapes.}
  \label{fig:optimtstsne}
\vspace{-0.15in}
\end{figure}

\noindent\textbf{Semantic Latent Space. }The latent space that we learn is semantic. We visualize the reconstructed full frequency coverage optimization process in the self-reconstruction during testing. The transformation from one latent code to another in Fig.~\ref{fig:optim} shows semantic shapes on the optimization path. Moreover, we can also manipulate embeddings in a semantic way like the plus and minus of embeddings for full frequency components in Fig.~\ref{fig:optimtadd}. We reduce the dimensions of all embeddings learned for low and full frequency components in each iteration using TSNE~\cite{vanDerMaaten2008} in Fig.~\ref{fig:optimtstsne}. We also see semantic structures like lines, each of which is formed by embeddings learned in all iterations (the order is mapped from light to dark color) on an optimization path. All optimization paths start from a similar point, goes quite similar in the beginning and become diverse at the end which corresponds to different shapes.

\section{Conclusion}
We introduce frequency consolidation priors to sharpen neural implicit functions. We successfully learn the priors from an established set containing training pairs with low frequency components and full frequency coverage. The learned priors can seamlessly work with our novel ways of recovering full frequency coverage from a low frequency observation, which significantly increases the generalization ability of the learned priors. We show that the learned priors can recover high frequency components from the low frequency observation, which sharpens the surfaces but also completes some missing structures. Our numerical and visual comparisons with the latest methods on widely used shape or scene datasets show that our priors can recover geometries with higher frequencies by sharpening low frequency SDF observation than the latest methods.

\bibliography{papers}

\end{document}